\documentclass[conference]{IEEEtran}
\usepackage{graphicx}
\usepackage{enumerate}
\usepackage{amssymb}
\usepackage{amsmath}
\usepackage{multirow}
\usepackage{booktabs}
\usepackage{float}

\begin{document}
	\IEEEoverridecommandlockouts
	
	\IEEEpubid{\makebox[\columnwidth]{978-1-4799-7560-0/15/\$31 \copyright 2015 IEEE \hfill} \hspace{\columnsep}\makebox[\columnwidth]{ }}
	
\title{Applying Interval Type-2 Fuzzy Rule Based Classifiers Through a Cluster-Based Class Representation}

\author{
\IEEEauthorblockN{J. Navarro, C. Wagner and U. Aickelin}
\IEEEauthorblockA{School of Computer Science\\University of Nottingham\\Nottingham, United Kingdom\\
Email: psxfjn@nottingham.ac.uk\\christian.wagner@nottingham.ac.uk\\uwe.aickelin@nottingham.ac.uk}}
\maketitle


\begin{abstract}
Fuzzy Rule-Based Classification Systems (FRBCSs) have the potential to provide so-called interpretable classifiers, i.e. classifiers which can be introspective, understood, validated and augmented by human experts by relying on fuzzy-set based rules. This paper builds on prior work for interval type-2 fuzzy set based FRBCs where the fuzzy sets and rules of the classifier are generated using an initial clustering stage. By introducing Subtractive Clustering in order to identify multiple cluster prototypes, the proposed approach has the potential to deliver improved classification performance while maintaining good interpretability, i.e. without resulting in an excessive number of rules.
The paper provides a detailed overview of the proposed FRBC framework, followed by a series of exploratory experiments on both linearly and non-linearly separable datasets, comparing results to existing rule-based and SVM approaches. Overall, initial results indicate that the approach enables comparable classification performance to non rule-based classifiers such as SVM, while often achieving this with a very small number of rules.
\end{abstract}

\IEEEpeerreviewmaketitle

\section{Introduction}

Fuzzy Rule-Based Classification Systems (FRBCSs) have been successfully applied as autonomous classifier and decision support systems in numerous classification problems since they provide a powerful platform to deal with uncertain, imprecise and noisy information while providing good interpretability in the form of IF-THEN rules (e.g.: \cite{herman2008design}, \cite{chiu1997extracting}, \cite{cordon1999proposal}, \cite{lopez2013hierarchical} and \cite{ishibuchi2005rule}).


FRBCs as well as Fuzzy Logic Systems can be classified by the type of Fuzzy Sets used, namely \textit{type}-1 and \textit{type}-2. In this context, performance improvements/advantages of the use of Interval type-2 Fuzzy Sets (IT2 FSs) and their applications over the type-1 counterpart have been found in several applications and fields, such as: IT2 FSs used in fuzzy clustering \cite{hwang2007uncertain}, fuzzy logic control to video streaming across IP networks \cite{jammeh2009interval}, fuzzy logic modelling and classification of video traffic using compressed data \cite{liang2001mpeg} and classification of imaginary hand movements in an EEG-based Brain-Computer Interface \cite{herman2006investigation}). These improvements have been attributed due to the additional degrees of freedom for uncertainty modelling in IT2 FSs and, in the case of classification problems, their capability to define/represent abstract classes.\\
There are many methods to generate fuzzy rules from known data, including heuristic approaches \cite{ishibuchi1992distributed}, genetic algorithms \cite{casillas2001genetic} \cite{ishibuchi1994construction} \cite{ishibuchi2001three}, neuro-fuzzy techniques \cite{mitra1995improving} \cite{nauck1997neuro}, data mining techniques \cite{de2005elicitation} \cite{hu2003elicitation}, and clustering methods \cite{roubos2003learning} \cite{chiu1997extracting}. In \cite{chiu1997extracting}, the use of an algorithm called Subtractive Clustering (based on determining a potential value as cluster center to each point in terms of its distance to the rest of the points in the space) helped to find rules for pattern classification by partitioning the input space into appropriate fuzzy regions for separation of classes.

\begin{figure} [H]
	\includegraphics[scale=0.45]{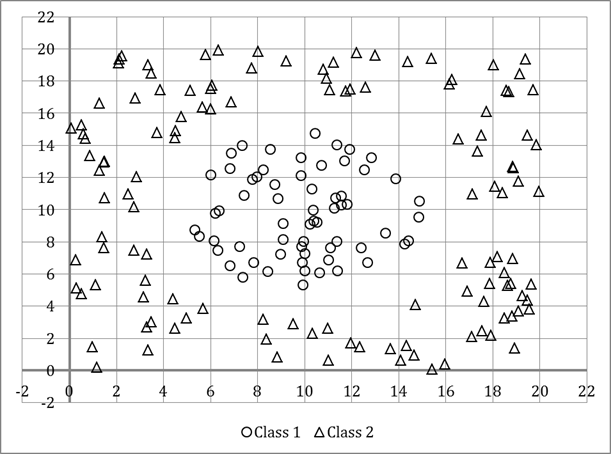}
	\centering
	\caption{{\fontsize{8pt}{1em} \selectfont An example of a non-linear classification problem between 2 classes in which one class is surrounded by a second class of different distribution and density.}}
	\label{fig:hat}
\end{figure}

The motivation of this paper builds on an a improvement to the IT2 FRBC model proposed in \cite{tang2011fuzzy} (based on rule generation method \cite{cordon1999proposal} derived from Wang and Mendel algorithm \cite{wang1992generating}) in which, the selection of more clusters for the representation of a single class is suggested but not explored. Building on this, this paper seeks to explore the fuzzy rules extraction method to estimate and select numbers and positions of representative cluster prototypes which together represent classes in linear and non-linear classification problems (such as depicted in the example in Fig. \ref{fig:hat}).


Nonlinear classification problems have been addressed through different approaches, such as establishing hierarchical connections based on the distance between objects (e.g. Decision Trees \cite{ittner1996discovery}, K-Nearest Neighbour \cite{min2009deep}), combining linear classifiers (e.g. Artificial Neural Networks \cite{khan2001classification}), learning a set of mappings from input-output relationships by changing the feature representation through a kernel (e.g. non-linear SVMs \cite{burges1998tutorial}), etc.

The overall aim in this paper is to generate FRBCs which provides good classification performance while maintaining good interpretability and thus, a low number of IF-THEN rules. In order to achieve our goal, we proposed the following developments to the FRBC model introduced in \cite{tang2011fuzzy}.
\begin{enumerate}
	\item While in \cite{tang2011fuzzy}, the initial centroids representing the classes are calculated as the mean of the training patterns, we proposed the use of SC to identify initial cluster prototypes.
	
	\item The resulting improvement must be able to provide superior performance in particular for non-linear problems, whereas the FRBC model proposed in \cite{tang2011fuzzy} is suitable for problems with circular distributions of clusters (of different size/density) but, it is not suitable for classification problems involving non-linearly separable classes.
\end{enumerate}
While SC provides the potential for such improved performance of the FRBC framework, in particular in non-linear classification problems, SC is highly sensitive to initial parameters which influence the number of clusters generated. As an increasing number of clusters results effectively in an increasing number of rules within the FRBC, we conduct an analysis on a number of publicly available and synthetic datasets, exploring the relationships between good interpretability (i.e. fewer rules) and good classification performance.



In Section II we present an introduction to interval type-2 fuzzy sets, Subtractive Clustering and general structure of Fuzzy Rule-Based Classification Systems. In Section III, we describe the approach used to represent a class through different clusters by taking SC as a method for finding potential representative clusters. In Section IV, the results of different numbers of clusters used to represent the classes in the FRBCs and comparisons against SVMs are being shown by considering 2 classic applications (Iris Plant and Wisconsin datasets) and two synthetic nonlinear classification problems. Conclusions and future work are presented in last section.

\section{Background}\label{background}
In order to ease the reader's understanding along the paper, we include a Table of symbols (see Table \ref{tab:symbols}) used often in most sections.
\begin{table}[h]
  \centering
  \caption{Table of symbols}
    \begin{tabular}{rp{4.5cm}}
    \toprule
    Symbol & Description \\
    \midrule
    $ \textbf{x} $     & Pattern \\
    $ n $     & number of training patterns \\
    $ r_{a} $    & The radius defining a neighbourhood in Subtractive Clustering \\
    $ j=1,…,M $ & Number of classes \\
    $ k=1,…,c $ & Number of rules/clusters \\
    $ l_{j} $    & Number of clusters related to class $ j $ \\
    $ r^{k}_{j} $   & Certainty degree \\
    $ \widetilde { A } $     & IT2 Fuzzy set \\
    $ P_{i} $    & Potential of $ x_{i} $ pattern \\
    $ C $     & Set of $ M $ classes \\
    $ f_{Q} $    & Quasiarithmetic mean operator \\
    $ m_{1} $    & Lower fuzzifier \\
    $ m_{2} $    & Upper fuzzifier \\
    \bottomrule
    \end{tabular}%
  \label{tab:symbols}%
\end{table}%

\subsection{Subtractive Clustering}\label{sec2:sub1}
Subtractive Clustering is a fast and robust method for estimating the number and location of cluster centers present in a collection of data points \cite{chiu1994fuzzy}. It has been widely used as a method for initial centroid selection in clustering methods such as FCM and Kohonen’s Self-Organizing Maps \cite{kohonen1982self} and shown to deliver good performance. It was proposed as an extension to the Yager and Filev's mountain method \cite{yager1994generation} in which each data point is considered as a potential cluster center to be accepted or rejected. In general terms it works as follows: consider a collection of $ n $ data points in which the potential as cluster center of the point $ x_{i} $ is defined initially as:

\begin{equation}
{ P }_{ i }=\sum _{ j=1 }^{ n }{ { e }^{ -\alpha { \left\| { x }_{ i }-{ x }_{ j } \right\|  }^{ 2 } } } 
\end{equation}
 where $ \alpha = 4/{ r }_{ a }^{ 2 }$ and $ r_{a} $ is a positive constant. Thus, the measure of the potential for a data point is a function of its distances to all other data points. A data point with many neighboring data points will have a high potential as cluster center. The constant $ r_{a} $ is effectively the radius defining a neighborhood and acts as a threshold to accept or refuse new cluster centers; data points outside this radius have little influence on the potential.
 
 After the potential of every data point has been computed, we select the data point with the highest potential as the first cluster center. Let $ { x }_{ 1 }^{ * } $ be the location of the first cluster center and $ { P }_{ 1 }^{ * } $ its potential value. Secondly, the potential of each data point $ x_{i} $ is revised using:
 \begin{equation}\label{eq:sc2}
{ P }_{ i } = { P }_{ i }-{ P }_{ 1 }^{ * }{ e }^{ -\beta { \left\| { x }_{ i }-{ x }_{ 1 }^{ * } \right\|  }^{ 2 } }
 \end{equation}
 where $ \beta = 4/{ r }_{ b }^{ 2 }$ and $ r_{b} $ is a positive constant. Thus, the potential value of each point is reduced by subtracting an amount of potential as a function of its distance from the first cluster center. Also, the data points near the first cluster center will have greatly reduced their potential, and therefore will unlikely be selected as the next cluster center. Commonly, the value of $ r_{b} $ is set to $ r_{b} = 1.25 r_{a} $.
 
 After the potential of all data points has been revised according to (\ref{eq:sc2}) we select the data point with the highest remaining potential as the second cluster center. We then further reduce the potential of each data point according to their distance to the second cluster center. In general, after the \textit{k}th cluster center has been obtained, the potential of each data point is revised similarly and, through an iterative algorithm, subsequent cluster centers are being accepted or rejected until a condition based on the threshold $ r_{a} $ is met. For further details on the algorithm can be found in \cite{chiu1994fuzzy}.

\subsection{Interval Type-2 Fuzzy Sets}
Fuzzy Set theory was introduced by Zadeh in 1965 \cite{zadeh1965fuzzy} and has been successfully applied in numerous fields in which uncertainty is involved. Type-2 Fuzzy Sets were introduced in 1975 \cite{zadeh1975concept} and their application has been shown to provide very good results in situations where lots of uncertainties are present \cite{mendel2003type}.

A T2 FS, denoted $ \widetilde { A }  $, characterized by a \textit{type-2} MF $ { \mu  }_{ \widetilde { A }  }\left( x,u \right)  $, where $ x\in X $ and $ u\in { J }_{ x }\subseteq \left[ 0,1 \right]  $, i.e.,

\begin{equation}
 \widetilde { A } =\left\{ \left( \left( x,u \right) ,{ \mu  }_{ \widetilde { A }  }\left( x,u \right)  \right) |\forall x\in X, \forall u\in { J }_{ x }\subseteq \left[ 0,1 \right]  \right\}
\end{equation}

If all $ { \mu  }_{ \widetilde { A }  }\left( x,u \right) = 1 $ then $ \widetilde { A }  $ is an Interval \textit{Type-2} FS (IT2 FS). 


Note that in IT2 FSs, since the third dimension (secondary grades) does not convey additional information, the Footprint of Uncertainty \cite{mendel2001uncertain} is sufficient to completely describe an IT2 FS. This simplification has proven to be useful to avoid the computational complexity of using general type-2 fuzzy sets (T2 FSs) \cite{mendel2001uncertain} and has been used in several fields, being pattern recognition one of them.

\subsection{Fuzzy Rule Based Classification Systems}\label{subsection3}
Pattern classification involves assigning a class $ C_{j} $ from a predefined class set $ C = { C_{1},..., C_{M} } $ to a pattern $ \textbf{x} $ in a feature space $ \textbf{x}\in { \mathbb{R} }^{ N }$. The purpose of a Classification System is to find a mapping $ D: { \mathbb{R} }^{ N }\rightarrow C $ optimal in the sense of a criterion that determines the classifier performance.

A FRBC consists of a Rule Base (RB) and a Fuzzy Reasoning Method (FRM) to derive class associations from the information provided by the set of rules fired by a pattern. 


\subsubsection{Fuzzy Rules Structure}
Considering a new pattern $ x =\left( { x }_{ 1 },...,{ x }_{ N } \right) $ and \textit{M} classes, three basic rule structures are identified in FRBCs according to \cite{cordon1999proposal}:

\begin{enumerate} [a)]
\item \textit{Fuzzy rules with a class in the consequent}: This type of the rule has the following structure:
\begin{center}
\textbf{IF} $ x_{ 1 } $ is $ { A }_{ 1 }^{ k } $ and ... and $ x_{N} $ is $ { A }_{ N }^{ k } $ \textbf{THEN} \textit{Y} is $ C^{k} $
\end{center}
where $ { x }_{ 1 },...,{ x }_{ N } $ are selected features for the classification problem, $ { A }_{ 1 }^{ k },..., { A }_{ N }^{ k }$ are linguistic labels modeled by fuzzy sets, and $ C^{k} $ is one of the predefined classes from the set $ C = { C_{1},..., C_{M} } $

\item \textit{Fuzzy rules with a class and a certainty degree in the consequent}: This type of the rule has the following structure:
\begin{center}
\textbf{IF} $ x_{ 1 } $ is $ { A }_{ 1 }^{ k } $ and ... and $ x_{N} $ is $ { A }_{ N }^{ k } $ \textbf{THEN} \textit{Y} is $ C^{k} $ with $ r^{k} $
\end{center}
where $ r^{k} \in [0,1]$ is the certainty degree of the \textit{k} rule. The value $ r^{k} $ can also be understood as a rule weight.

\item \textit{Fuzzy rules with certainty degree for all classes in the consequent}: This type of the rule has the following structure:
\begin{center}
\textbf{IF} $ x_{ 1 } $ is $ { A }_{ 1 }^{ k } $ and ... and $ x_{N} $ is $ { A }_{ N }^{ k } $ \textbf{THEN} $ ({r}_{1}^{k},...,{r}_{M}^{k}) $
\end{center}
where $ r_{j}^{k} \in [0,1]$ is the certainty degree for rule \textit{k} to predict the class $ C_{j} $ for a pattern belonging to the fuzzy region represented by the antecedent of the rule.
\end{enumerate}

\subsubsection{Fuzzy Reasoning Method}
Given an input pattern $ \textbf{x} $, conclusions can be derived using a FRM based on the RB. The inference process of a FRM for IT2 FSs is summarized below (this is based on the type-1 formulation given in \cite{tang2011fuzzy}):

\begin{enumerate}
\item \textit{Matching degree}. The strength of activation of the antecedent for rule $ k $ in the Rule Base (RB) with the pattern $ \textbf{x} =(x_{1},...,x_{N})$ is calculated by using a function $ T $ which is a \textit{t}-norm (commonly minimum \textit{t}-norm).
\begin{equation} \label{eq:matchingDegree}
{ R }^{ k }(\textbf{x})=T({ \mu  }_{ { A }_{ 1 }^{ k } }({ x }_{ 1 }),...,{ \mu  }_{ { A }_{ N }^{ k } }({ x }_{ N }))
\end{equation}

\item \textit{Association degree}. The association degree of the pattern $ \textbf{x} $ with the \textit{M} classes is computed in each rule by calculating the product:
\begin{equation} \label{eq:associationDegree}
{ b }_{ j }^{ k }=\left( { R }^{ k }\left( { \textbf{x} } \right) \cdot { r }_{ j }^{ k } \right),
\end{equation}
where $ r_{j}^{ k } $ stands for the certainty degree provided in rule $ k $ for the class $ j $.
\item \textit{Weighting function}. To weight the obtained association degrees through a function \textit{g}. Commonly, a weighting function boosts big output values and suppresses small output values.

\item \textit{Pattern classification soundness degree for all classes}. An aggregation operator $ f $ (for example the Quasiarithmetic mean operator, see below) is used to combine (the positive degrees of association calculated in the previous step) into a single value $ Y_{j} $, which is the soundness degree of pattern $ \textbf{x} $ with class $ C_{j} $.
\begin{equation} \label{eq:soundnessDegree}
{ Y }_{ j }=f\left( { b }_{ j }^{ k },k=1,...,c \right) 
\end{equation}
The soundness degree $ Y_{j} $ expresses in one value, to what extent the information of the rules have contributed for the classification into a given class $ j $.

\item \textit{Classification}. A decision function \textit{h} is applied over the soundness degree for all classes. This function determines the class $ C_{j} $ corresponding to the maximum value obtained. This is $ C_{j} = h(Y_{1},...,Y_{M}) $ such that
\begin{equation}\label{eq:classification}
  Y_{j} =\max _{ j=1,...,M }{ { Y }_{ j } }
\end{equation}
\end{enumerate}


\subsubsection{Aggregation Operators}
Aggregation Operators are important since their use allows a FRM to consider the information given by all the rules compatible with an input pattern and their selection is dependant of the problem. For these reasons, several Aggregation Operators have been proposed such as the ones described in \cite{cordon1999proposal}. Below, we describe the operator used in the original FRM from \cite{tang2011fuzzy} also used in our experiments.

Let $ \left( { a }_{ 1 },...,{ a }_{ s } \right) $ represent the association degrees $ { b }_{ j }^{ k }$ of $ c $ rules such that $ { b }_{ j }^{ k }>0 $ and $ k=1,...,c $ for a given pattern and one class $ j $:
\begin{equation}\label{eq:Quasi}
{ f }_{ Q }^{ j }\left( { a }_{ 1 },...,{ a }_{ s } \right) =\left[ \frac { 1 }{ s } \sum _{ l=1 }^{ s }{ \left( { a }_{ l } \right) ^{ p } }  \right]^{ \frac { 1 }{ p }  } 
\end{equation}
where $ p\in \mathbb{R} $ and $ { f }_{ Q }^{ j } $ stands for the Quasiarithmetic mean operator applied to the non-zero degrees of association to a given class $ C_{j} $ provided by the weighting function. The behaviour of this operator produces an aggregated value of degree of association between the minimum and the maximum and is determined by the selection of $ p $ such that:
\begin{itemize}
	\item If $ p\rightarrow -\infty  $, $ { f }_{ Q } \rightarrow $ min,
	\item If $ p\rightarrow +\infty  $, $ { f }_{ Q } \rightarrow $ max.
\end{itemize}
For more detail about its properties see \cite{dyckhoff1984generalized}.

The implementation of aggregation operators in the FRM provides an inference process capable of using a good combination of the rules information to define class membership. In the next section, a description of our proposed FRBC through a Cluster-Based Representation is presented.

\section{FRBC Construction with Cluster-Based Class Representation}
In order to explain both the generation of the rules from training data and classification process, we explain all steps as shown in Fig. \ref{fig:RuleExtraction}:

\begin{figure} [H]
	\includegraphics[scale=0.45]{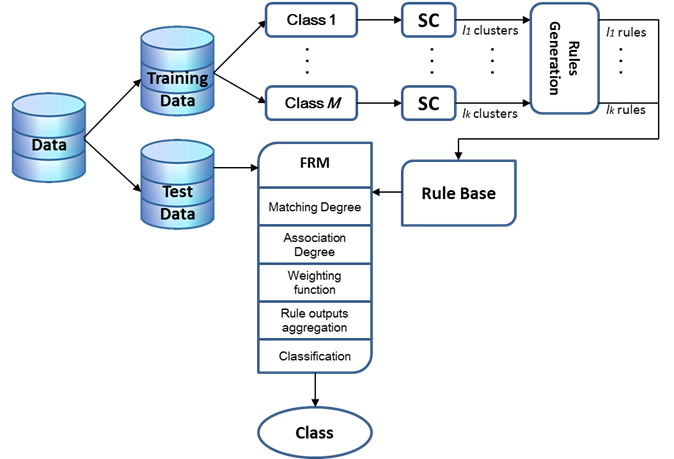}
	\centering
	\caption{{\fontsize{8pt}{1em} \selectfont Rule extraction and classification process}}
	\label{fig:RuleExtraction}
\end{figure}

\subsection{Rule Base Construction}
Consider a given set of $ n $ patterns for training a FRBC with \textit{M} classes.

\begin{enumerate}
	\item Separate training dataset per class and perform SC in each training subset.
	\item Let the number of clusters found by SC for class $ C_{j} $ be $ l_{j} $, set the total number of clusters as $ c=\sum _{ j=1 }^{ M }{l_{j}} $.
	
	\item Calculate the matrix of distances from each $ i $ training pattern to each $ k $ cluster.
	
	\item By using two fuzzifiers $ m_{1} $ and $ m_{2} $, calculate the membership degrees of each training pattern $ \textbf{x} $ to a cluster \textit{k} such as shown in (\ref{eq:MF1}) and (\ref{eq:MF2}):
	
	\begin{equation}\label{eq:MF1}
	{ \mu  }_{ k }^{ m_{ 1 } }=\frac { 1 }{ \sum _{ q=1 }^{ c }{ { \left( { \frac { { d }_{ k } }{ { d }_{ q } }  } \right)  }^{ \frac { 2 }{ \left( { m }_{ 1 }-1 \right)  }  } }  }  
	\end{equation}
	
	\begin{equation}\label{eq:MF2}
	{ \mu  }_{ k }^{ m_{ 2 } }=\frac { 1 }{ \sum _{ q=1 }^{ c }{ { \left( { \frac { { d }_{ k } }{ { d }_{ q } }  } \right)  }^{ \frac { 2 }{ \left( { m }_{ 2 }-1 \right)  }  } }  } ,
	\end{equation}
	
	where $ d_{k} $ is the distance to the \textit{k}th cluster prototype, and \textit{k} = 1,..., \textit{c}. This way, the uncertainty associated with the size and density of the clusters representing a class is being managed through the fuzzifier \textit{m}, such as shown in (\ref{eq:upMF}) for the upper bound of the membership:
	
	\begin{equation}\label{eq:upMF}
	\overline { \mu  } _{ k }=\max { \left( { \mu  }_{ k }^{ m_{ 1 } },{ \mu  }_{ k }^{ m_{ 2 } } \right)  }  ,
	\end{equation}
	whereas for the lower bound we consider (\ref{eq:lwMF}):
	\begin{equation}\label{eq:lwMF}
	\underline { \mu  } _{ k }=\min { \left( { \mu  }_{ k }^{ m_{ 1 } },{ \mu  }_{ k }^{ m_{ 2 } } \right)  } 
	\end{equation}
	
	Thus, a footprint of uncertainty is created by using the highest and lowest primary memberships of a pattern $ \textbf{x} $ to a cluster \textit{k}.
	
	\item Generate the \textit{c} IT2 FSs and their respective MFs according to (\ref{eq:upMF}) and (\ref{eq:lwMF}) such that the MF of the \textit{k}th IT2 FS named $ \widetilde { A }_{ k } $ is denoted by
	\begin{equation}
	 \mu_{{ \widetilde { A }  }_{ k }}=\left[ { \underline { \mu  } _{ k }\left( \textbf{x} \right)  },{ \overline { \mu  } _{ k }\left( \textbf{x} \right)  } \right]    
	\end{equation}

	\item Construct a type-2 fuzzy rule base with \textit{c} rules, $ \widetilde { A }_{ k } $ being the single antecedent in each rule. The certainty degree is defined as 
	\begin{equation}
	{ r }_{ j }^{ k }=\frac { \sum _{ c\left( { \textbf{x} }_{ i } \right) ={ c }_{ j } }{ { U }_{ k }\left( { \textbf{x} }_{ i } \right)  }  }{ \sum _{ i=1 }^{ n }{ { U }_{ k }\left( { \textbf{x} }_{ i } \right)  }  }   
	\end{equation} 
	where \textit{j} = 1,...,\textit{M}, $ c\left( { \textbf{x} }_{ i } \right)  $ denotes the class label of training pattern $ { \textbf{x} }_{ i } $, \textit{n} is the number of training patterns and
	\begin{equation}
{ U }_{ k }\left( { \textbf{x} }_{ i } \right) =\frac { { \underline { \mu  } _{ k }\left( \textbf{x}_{i} \right)  }+{ \overline { \mu  } _{ k }\left( \textbf{x}_{i} \right)  } }{ 2 }  
	\end{equation}
\end{enumerate}

\begin{figure} [H]
	\includegraphics[scale=0.45]{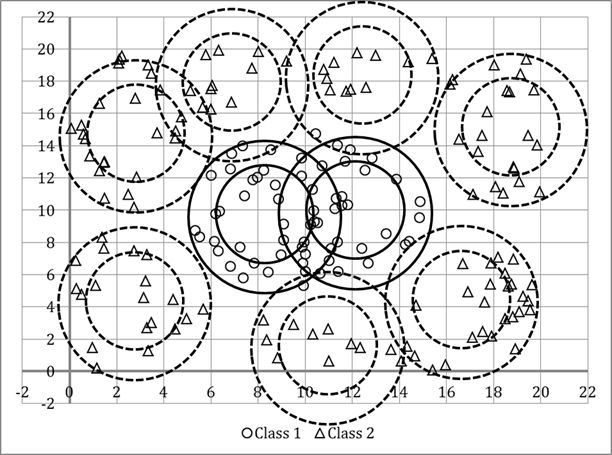}
	\centering
	\caption{{\fontsize{8pt}{1em} \selectfont A possible cluster based representation solution for a non-linear classification problem. The continuum circles stand for the clusters representing the Class 1, whereas the dashed circles stand for the Class 2}}
	\label{fig:surrounded}
\end{figure}

Summarizing the methodology to construct the Rule Base, our proposed construction of the FRBC is performed in one single-pass by splitting the training data per class and applying SC subsequently in order to find representative clusters prototypes of a single class such as illustrated in Fig. \ref{fig:surrounded} where, non-circular dispersions of points are approximated by several circular clusters with uncertain size/density.

Clearly, the resulting clusters do not yet provide a useful answer in terms of classifying the data into the expected number of classes. The final stage of the process is addressed in the following section.

\subsection{Inference process}
Once the construction of the rules is performed, we have a rule base so further patterns can be used to test it by using the FRM as inference method for classification. Given a pattern $ \textbf{x} $ to classify we follow these steps:

\begin{enumerate}
\item Calculate its matching degree in each rule by determining its interval of membership to each cluster $ k $:
\begin{equation} \label{IMatchingDegree}
{ R }^{ k }({ \textbf{x} })=\mu _{ { \widetilde { A }  }_{ k } }=\left[ { \underline { \mu  } _{ { A }_{ k } }\left( { \textbf{x} } \right)  },{ \overline { \mu  } _{ { A }_{ k } }\left( { \textbf{x} } \right)  } \right]    
\end{equation}
Note that in these rules there is just one antecedent so, considering (\ref{eq:matchingDegree}), the matching degree is the degree of membership in $ \widetilde { A }_{ k } $.

\item Considering (\ref{eq:associationDegree}) and (\ref{IMatchingDegree}), calculate its association degree in the interval by using product\\
\begin{equation} \label{Iassociation}
{ b }_{ j }^{ k }=\left[ { b }_{ jl }^{ k },{ b }_{ jr }^{ k } \right] =\left[ \left( { \underline { \mu  } _{ { A }_{ k } }\left( { \textbf{x} } \right) \cdot { r }_{ j }^{ k } } \right) ,\left( \overline { \mu  } _{ { A }_{ k } }\left( { \textbf{x} } \right) \cdot { r }_{ j }^{ k } \right)  \right] 
\end{equation}
where $ { b }_{ j }^{ k }  $ is the association degree to the $ j $ class under the \textit{k}th rule.\\

\item In this FRM, the weighting function $ g $ is $ g(x)=x $ such as in \cite{tang2011fuzzy}.\\

\item Considering (\ref{eq:soundnessDegree}) and (\ref{Iassociation}), calculate the soundness degree $ Y_{j} $ of pattern $ \textbf{x} $ with class $ C_{j} $ by
\begin{equation} \label{Isoundness}
{ Y }_{ j }=\left[ { Y }_{ jl },{ Y }_{ jr } \right] =\left[ { Y }_{ jl }=f\left( { b }_{ jl }^{ k } \right) ,{ Y }_{ jr }=f\left( { b }_{ jr }^{ k } \right)  \right] 
\end{equation} 
where $ k=1,...,C $ and $ f $ stands for the Quasiarithmetic mean aggregation operator defined in (\ref{eq:Quasi}).\\

\item Finally, the decision function $ h $ is applied over the soundness degree interval in all classes. This is performed by considering (\ref{eq:classification}) and (\ref{Isoundness}) so, assign the class $ C_{j} $ to the pattern $ \textbf{x} $ such that the function $ h(Y_{1},...,Y_{M}) $ determines the class corresponding to the maximum value obtained. This is:
\begin{equation}
{ Y }_{ j }=\max _{ j=1,...,M }{ \left( \frac { { Y }_{ jl }+{ Y }_{ jr } }{ 2 }  \right)  } 
\end{equation}
Note that, contrary to (\ref{eq:classification}), here we are considering the average of the soundness degree bounds rather than a single value in order to determine the class to be assigned.
\end{enumerate}

\section{Results}\label{results}
In this section, the utility of IT2 FRBCs is demonstrated on four datasets by comparing their results against the ones of a SVM using a radial basis function. We used $ m_{1}=1.5 $ and $ m_{2}=2.5 $ as fuzziness parameters for IT2 FSs. These values were chosen based on the Pal and Bezdek \cite{pal1995cluster} study which suggests that the best choice for \textit{m} (based on the performance of some cluster validity indices) is commonly in the interval [1.5, 2.5]. Also in SC applications, normalization to $ [0,1] $ along with a $ r_{a} $ value between [0.4, 0.6] is recommended to get reasonable results but, in this application, different values of $ r_{a} $ are being explored since different numbers of clusters prototypes were found in the data and consequently the number of rules. 

\subsection{Non-linear classification problems}
\subsubsection{Circular surrounding}
Our first experiment is performed with the synthetic data set shown in Fig. \ref{fig:hat}, where there are 186 patterns with two features generated randomly in the interval [0, 20]. Then we proceed to label the patterns according to their Euclidean distance to the point (10, 10) so if the distance of a pattern $ x_{i} $ is greater than 7, then the label assigned is \textit{class} 2, otherwise if its distance is lower than 5 it is associated to \textit{class} 1. Thus, we generated a circular distribution of the class 1 with 63 patterns and, on the other hand, we generated 123 patterns for the class 2 surrounding the former class with different sizes and densities. In order to create the FRBC we shuffled and divided the data into 2 sets,  50\%  to construct the model and the rest for test. This process was repeated over 32 experiments per each $ r_{a} $ value and the SVM as well. In Table \ref{tab:nonlinear1}, we show the results of these experiments, where the number of clusters found in the 32 runs is reported as an interval for the minimum and maximum numbers. When SC was not used, we chose the mean of the training patterns in each class as the initial prototype for each cluster (such as made in \cite{tang2011fuzzy}).

\begin{table}[htbp]
  \centering
  \caption{Cluster-based class representation in a synthetic Non-Linear problem}
    \begin{tabular}{|c|cccccc|}
    \toprule
          & \textbf{$ r_{a} $} & \textbf{Clusters/rules} & \textbf{Best} & \textbf{Average} & \textbf{Worst} & \textbf{$ \sigma  $} \\
    \midrule
        \multirow{6}[2]{*}{ FRBC } & none  & 2     & 74.19 & 66.30 & 58.06 & 3.52 \\
          & 0.2   & [19,25] & \textbf{100} & \textbf{99.16} & 95.70 & 1.10 \\
          & 0.3   & [10,14] & \textbf{100} & 96.94 & 90.32 & 2.39 \\
          & 0.4   & [8,10] & \textbf{100} & 95.23 & 90.32 & 2.51 \\
          & 0.5   & [6,8] & 98.92 & 94.72 & 87.10 & 2.80 \\
          & 0.6   & [6,7] & 98.92 & 93.75 & 89.25 & 2.67 \\\hline
    SVM   &       &       & \textbf{100}   & 97.85 & 93.55 & 1.68 \\
    \bottomrule
    \end{tabular}
  \label{tab:nonlinear1}
\end{table}

As can be seen, the results of using more than one cluster for representing a single class were superior over the use of a single cluster but note that, as we increased the $ r_{a} $ threshold and consequently the number of clusters/rules were reduced (see Section \ref{sec2:sub1}), the average performance of the FRBCs was reduced. Nevertheless, in all cases there was a significant improvement over the original FRBC model (one cluster to represent each class), which in all cases misclassified all patterns of \textit{class} 1 to \textit{class} 2. 


Regarding the comparisons with SVM results, the average of the 32 experiments was found comparable to the constructed FRBCs, outperformed only by the FRBC where $ r_{a} = 0.2$.

\subsubsection{Irregular distribution}
\begin{figure}
	\includegraphics[scale=0.45]{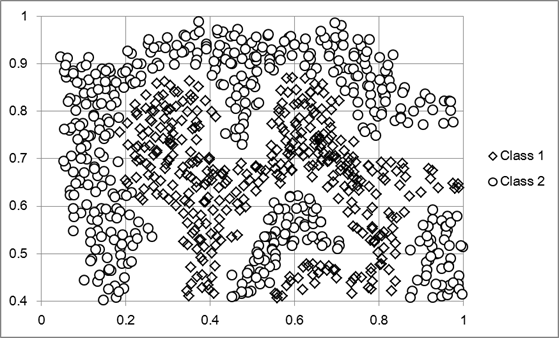}
	\centering
	\caption{{\fontsize{8pt}{1em} \selectfont Two non-linearly separable classes}}
	\label{fig:ex8a}
\end{figure}

Our second experiment is performed with the pattern set shown in Fig. \ref{fig:ex8a} where there are 863 patterns with two features. There are 480 and 383 patterns respectively for class 1 and class 2 over an irregular distribution surrounding the latter class with different sizes and densities.

  Similarly as in the previous application, we shuffled the data and divided it into 2 sets in which 50\% was used to construct the model and the rest for testing. This process was repeated 32 times. In Table \ref{tab:nonlinear2}, we show the results of these experiments, where the number of clusters found for all the classes in the 32 iterations is being reported as an interval for the lower and bigger amounts.
\begin{table}[htbp]
  \centering
  \caption{Cluster-based class representation in a Synthetic Non-Linear problem}
    \begin{tabular}{|c|cccccc|}
    \toprule
          & \textbf{$ r_{a} $} & \textbf{Clusters/rules} & \textbf{Best} & \textbf{Average} & \textbf{Worst} & \textbf{$ \sigma  $} \\
    \midrule
    \multirow{3}[2]{*}{ FRBC } & none  & 2     & 62.96 & 56.32 & 52.08 & 2.33 \\
    & 0.2   & [23,30] & 98.61 & 95.43 & 91.20 & 1.51 \\
          & 0.4   & [10,13] & 91.44 & 85.63 & 82.41 & 1.84 \\
          & 0.6   & [5,9] & 80.09 & 71.92 & 58.80 & 4.39 \\\hline
    SVM   &       &       & \textbf{99.76} & \textbf{98.57} & 96.98 & 0.75 \\
    \bottomrule
    \end{tabular}%
  \label{tab:nonlinear2}%
\end{table}%
Similarly to the previous synthetic experiment, in all cases there was a significant improvement over the original FRBC model which uses a single prototype per class given that, by looking at the confusion matrices, there were misclassified patterns from \textit{class} 2 to \textit{class} 1. This shows that the results of the use of more than one cluster for the representation of a single class were superior to the use of a single prototype as expected.

Additionally we also found a negative correlation between the $ r_{a} $ value and the number of rules, as well as a positive correlation between the average performance of the FRBCs and the size of the rule base. These correlations, seem to indicate a high importance in the selection of the $ r_{a} $ value in order to control the sensitivity of the FRBC to accept reference points.  Regarding the comparisons with SVMs results, the 32 experiments were found superior to the constructed FRBCs, with the FRBC with $ r_{a}=0.2 $ the closest in average. 

\subsection{Application to Iris Plant Benchmark}
For our next experiment, we used the Iris Plant dataset from the UCI repository of machine learning databases. The Iris dataset is composed of 150 4-dimensional patterns uniformly distributed among three classes. For these experiments we performed the same methodology used in the previous experiments by dividing the dataset into 2 subsets, 50\% for training and 50\% for testing. Note that for comparison purposes, we followed the approach in \cite{tang2011fuzzy} as closely as possible. Our FRBCs results are shown in Table \ref{tab:iris} and also, we include the results reported in the original approach \cite{tang2011fuzzy}.

\begin{table}[htbp]
  \centering
  \caption{Cluster-based class representation in Iris Plant dataset}
    \begin{tabular}{|c|cccccc|}
    \toprule
          & \textbf{$ r_{a} $} & \textbf{Clusters/Rules} & \textbf{Best} & \textbf{Average} & \textbf{Worst} & \textbf{$ \sigma $} \\
    \midrule
    \multirow{5}[2]{*}{FRBC} & none  & 3     & 98.67 & 92.21 & 85.33 & 3.06 \\
		    & 0.2   & [21,32] & \textbf{100} & 94.46 & 89.33 & 2.38 \\
          & 0.3   & [8,16] & 98.67 & 95.08 & 90.67 & 2.07 \\
          & 0.4   & [6,10] & 98.67 & 94.63 & 85.33 & 3.01 \\
          & 0.5   & [6,9] & 98.67 & 94.04 & 86.67 & 2.85 \\
          & 0.6   & 6     & \textbf{100} & 93.54 & 85.33 & 3.09 \\\hline
          Tang \cite{tang2011fuzzy} &       & 3     & 97.33 & 88.80 & 72.00 &  \\\hline
    SVM   &       &       & \textbf{100}   & \textbf{96.95} & 92.00 & 1.98 \\
    \bottomrule
    \end{tabular}
  \label{tab:iris}
\end{table}%

As in previous experiments, some improvements were reached by using more than one cluster for class representation although in this case were subtle. By comparing with the results reported in \cite{tang2011fuzzy}, our cluster-based representation seems to improve the classification accuracy. We attributed this improvement to the use of different fuzziness parameters, since in the original paper they were chosen as: $ m_{1}=2 $ and $ m_{2}=5 $ and also, they refined the cluster prototypes by using IT2 FCM before constructing the Rule Base.

\subsection{Application to WBCD Benchmark}
Finally, we performed experiments in the same fashion with the Wisconsin breast cancer diagnosis (WBCD) database which is the result of the efforts made at the University Of Wisconsin Hospital for accurately diagnose breast masses based solely on an Fine Needle Aspiration (FNA) test. This dataset contains a total of 699 clinical instances, with 458 benign and 241 malignant cases. Each instance has 9 attributes but, 16 of the samples are each missing one of the nine attributes so in common practice, those instances are eliminated as in this experiments as well. In Table \ref{tab:wisconsin} we show these experiments results.

\begin{table}[htbp]
  \centering
  \caption{Cluster-based class representation in Wisconsin dataset}
    \begin{tabular}{|c|cccccc|}
    \toprule
          & \textbf{$ r_{a} $} & \textbf{Clusters/rules} & \textbf{Best} & \textbf{Average} & \textbf{Worst} & \textbf{$ \sigma $} \\
    \midrule
    \multirow{4}[1]{*}{FRBC} & none  & 2     & 98.25 & \textbf{96.54} & 94.44 & 0.79 \\
		  & 0.2   & [112,132] & 97.08 & 95.87 & 94.15 & 0.64 \\
          & 0.4   & [107,132] & 97.08 & 91.51 & 85.38 & 2.87 \\
          & 0.6   & [93,121] & 95.32 & 92.35 & 88.30 & 1.69 \\
          & 1     & [12,35] & 97.66 & 95.92 & 93.57 & 0.90 \\
          & 1.5   & [4,5] & \textbf{98.54} & 96.31 & 94.44 & 1.07 \\\hline
    SVM   &       &       & 96.48 & 95.30 & 93.25 & 0.85 \\
    \bottomrule
    \end{tabular}%
  \label{tab:wisconsin}%
\end{table}%
Contrary to findings in previous experiments, the recommended interval of $ r_{a} $ found several clusters in this dataset which seem to overfit the generated FRBC models, i.e. the use of more rules resulted counterproductive in most of the experiments although, non clear improvements were reached over the results of the FRBCs in which we did not perform SC despite the number of rules constructed. Additionally, by comparing the results of the generated FRBCs against the SVM results we found a slight difference favouring our approach.


\section{Discussion}
In the FRBC model presented, we explored different numbers of rules while seeking for reasonably good performance instead of focusing on optimizing MF parameters such as in other FRBCs applications (e.g. \cite{chiu1997extracting}). This exploration was addressed by considering different $ r_{a} $ values for SC given that, as mentioned in Section \ref{sec2:sub1}, the parameter $ r_{a} $ is crucial to determine the sensitivity of the algorithm to accept or refuse cluster centroids. Consequently, we found that $ r_{a} $ acts as a threshold to control the number of rules of the FRBC which can increase the computational cost related to the number of clusters/MFs to analyse and affect the interpretability. Also, we found that the relation between number of rules and performance is not always as expected since we observed that, in some cases, a relatively big number of rules does not necessarily help to improve the FRBC accuracy and affects the interpretability of the FRBCS (according to the discussion about considerations for interpretability in \cite{jin2000fuzzy}).


Another consideration for interpretability is the \textit{number of antecedents} in the rules. In this approach one rule is created for each cluster prototype found in SC while handling different certainty degrees, as shown in (\ref{eq:rules}) and determining the membership degrees of the training patterns to different clusters. Thus, the created rules keep the structure:
\begin{equation}\label{eq:rules}
{ R }_{ k }:\quad IF\quad x\quad is\quad { \widetilde { A }  }_{ k },\quad THEN\quad \left( { r }_{ 1 }^{ k },...,{ r }_{ M }^{ k } \right), 
\end{equation}
Thus, points of reference (class prototypes) are being used to measure the similarity of input patterns to the antecedents representing the points of reference within the feature space. Using type-2 FSs provides the potential to capture more complex mappings without adding additional rules, a feature which we seek to explore further in the context of FRBCs in the future.

Another important consideration for interpretability is related to the distribution of the membership functions and their \textit{distinguishability}. Here, as consequence of using SC to create the rules, the antecedent's membership functions are distinguishable from the rest in the pattern feature space. Also, the Rule-Base is \textit{consistent} (i.e. non-contradictory rules) since it is generated from the centroids provided by SC which follow certain separability and compactness (dependant on the threshold $ r_{a} $) so, rule antecedents related to different classes will be reasonably different.

A number of remaining challenges have been identified: the potential for overfitting due to a large number of cluster prototypes - as identified in the case of the Wisconsin dataset. Further, the fundamental relationship of performance and interpretability and how to automatically balance them (which includes measuring the level of interpretability at runtime) remains to be addressed. In other words, as remarked previously, an appropriate selection of the $ r_{a} $ value for SC is required to reach both characteristics of a FRBC: interpretability and good performance. 

Still, even in the face of these challenges, the results show that the relative simplicity of the FRBCs through this cluster-based class representation can help FRBCs to become candidates for both, linear and non-linear classification problems in which the advantages of FRBCs (e.g. interpretability of IF-THEN rules) is valuable.

\section{Conclusion}
Our main contribution in this work consists in showing the potential of a cluster-based representation in linear and non-linear classification problems while avoiding an space transformation of input data such as required in our reference method (SVM). This was done by starting from an improvement to the IT2 FRBC model described in \cite{tang2011fuzzy}. This improvement was reached by using distributed clusters of uncertain size due to the use of two fuzziness parameters in order to create a foot of uncertainty represented by an interval. This manner, the initial representation in \cite{tang2011fuzzy} which (during the experiments) misclassified all patterns of one class to another in the first two datasets of Section \ref{results}, was changed for a more appropriate representation.

We implemented Subtractive Clustering along with a FRBC managing IT2 FSs and we found that, the appropriate selection of value $ r_{a} $ is highly dependent of the size/density of the data. We also could determine that, in classification problems, a large number of rules do not necessarily helps to improve the accuracy of the FRBC.

We have presented four experiments while comparing to SVMs and focusing mainly in the potential of cluster-based representation to improve results in different scenarios such as nonlinear classification problems.




As part of future work, we aim to develop a methodology to balance classification performance and interpretability based on application-specific criteria.
Further, we will pursue more extensive comparisons with other techniques including other rule based approaches (e.g., ANFIS \cite{jang1991fuzzy} \cite{jang1993anfis}) and decision tree based classifiers.

\bibliographystyle{IEEEtran}
\bibliography{IEEEabrv,mybibfile}

\begin{thebibliography}{10}
\providecommand{\url}[1]{#1}
\csname url@samestyle\endcsname
\providecommand{\newblock}{\relax}
\providecommand{\bibinfo}[2]{#2}
\providecommand{\BIBentrySTDinterwordspacing}{\spaceskip=0pt\relax}
\providecommand{\BIBentryALTinterwordstretchfactor}{4}
\providecommand{\BIBentryALTinterwordspacing}{\spaceskip=\fontdimen2\font plus
\BIBentryALTinterwordstretchfactor\fontdimen3\font minus
  \fontdimen4\font\relax}
\providecommand{\BIBforeignlanguage}[2]{{%
\expandafter\ifx\csname l@#1\endcsname\relax
\typeout{** WARNING: IEEEtran.bst: No hyphenation pattern has been}%
\typeout{** loaded for the language `#1'. Using the pattern for}%
\typeout{** the default language instead.}%
\else
\language=\csname l@#1\endcsname
\fi
#2}}
\providecommand{\BIBdecl}{\relax}
\BIBdecl

\bibitem{herman2008design}
P.~Herman, G.~Prasad, and T.~M. McGinnity, ``Design and on-line evaluation of
  type-2 fuzzy logic system-based framework for handling uncertainties in bci
  classification,'' in \emph{Engineering in Medicine and Biology Society, 2008.
  EMBS 2008. 30th Annual International Conference of the IEEE}.\hskip 1em plus
  0.5em minus 0.4em\relax IEEE, 2008, pp. 4242--4245.

\bibitem{chiu1997extracting}
S.~Chiu, ``Extracting fuzzy rules from data for function approximation and
  pattern classification,'' \emph{Fuzzy Information Engineering: A Guided Tour
  of Applications.}, 1997.

\bibitem{cordon1999proposal}
O.~Cord{\'o}n, M.~J. del Jesus, and F.~Herrera, ``A proposal on reasoning
  methods in fuzzy rule-based classification systems,'' \emph{International
  Journal of Approximate Reasoning}, vol.~20, no.~1, pp. 21--45, 1999.

\bibitem{lopez2013hierarchical}
V.~L{\'o}pez, A.~Fern{\'a}ndez, M.~J. Del~Jesus, and F.~Herrera, ``A
  hierarchical genetic fuzzy system based on genetic programming for addressing
  classification with highly imbalanced and borderline data-sets,''
  \emph{Knowledge-Based Systems}, vol.~38, pp. 85--104, 2013.

\bibitem{ishibuchi2005rule}
H.~Ishibuchi and T.~Yamamoto, ``Rule weight specification in fuzzy rule-based
  classification systems,'' \emph{Fuzzy Systems, IEEE Transactions on},
  vol.~13, no.~4, pp. 428--435, 2005.

\bibitem{hwang2007uncertain}
C.~Hwang and F.~C.-H. Rhee, ``Uncertain fuzzy clustering: interval type-2 fuzzy
  approach to c-means,'' \emph{Fuzzy Systems, IEEE Transactions on}, vol.~15,
  no.~1, pp. 107--120, 2007.

\bibitem{jammeh2009interval}
E.~Jammeh, M.~Fleury, C.~Wagner, H.~Hagras, M.~Ghanbari \emph{et~al.},
  ``Interval type-2 fuzzy logic congestion control for video streaming across
  ip networks,'' \emph{Fuzzy Systems, IEEE Transactions on}, vol.~17, no.~5,
  pp. 1123--1142, 2009.

\bibitem{liang2001mpeg}
Q.~Liang and J.~M. Mendel, ``Mpeg vbr video traffic modeling and classification
  using fuzzy technique,'' \emph{Fuzzy Systems, IEEE Transactions on}, vol.~9,
  no.~1, pp. 183--193, 2001.

\bibitem{herman2006investigation}
P.~Herman, G.~Prasad, and T.~McGinnity, ``Investigation of the type-2 fuzzy
  logic approach to classification in an eeg-based brain-computer interface,''
  in \emph{Engineering in Medicine and Biology Society, 2005. IEEE-EMBS 2005.
  27th Annual International Conference of the}.\hskip 1em plus 0.5em minus
  0.4em\relax IEEE, 2006, pp. 5354--5357.

\bibitem{ishibuchi1992distributed}
H.~Ishibuchi, K.~Nozaki, and H.~Tanaka, ``Distributed representation of fuzzy
  rules and its application to pattern classification,'' \emph{Fuzzy sets and
  systems}, vol.~52, no.~1, pp. 21--32, 1992.

\bibitem{casillas2001genetic}
J.~Casillas, O.~Cord{\'o}n, M.~J. Del~Jesus, and F.~Herrera, ``Genetic feature
  selection in a fuzzy rule-based classification system learning process for
  high-dimensional problems,'' \emph{Information Sciences}, vol. 136, no.~1,
  pp. 135--157, 2001.

\bibitem{ishibuchi1994construction}
H.~Ishibuchi, K.~Nozaki, N.~Yamamoto, and H.~Tanaka, ``Construction of fuzzy
  classification systems with rectangular fuzzy rules using genetic
  algorithms,'' \emph{Fuzzy sets and systems}, vol.~65, no.~2, pp. 237--253,
  1994.

\bibitem{ishibuchi2001three}
H.~Ishibuchi, T.~Nakashima, and T.~Murata, ``Three-objective genetics-based
  machine learning for linguistic rule extraction,'' \emph{Information
  Sciences}, vol. 136, no.~1, pp. 109--133, 2001.

\bibitem{mitra1995improving}
S.~Mitra and L.~I. Kuncheva, ``Improving classification performance using fuzzy
  mlp and two-level selective partitioning of the feature space,'' \emph{Fuzzy
  Sets and Systems}, vol.~70, no.~1, pp. 1--13, 1995.

\bibitem{nauck1997neuro}
D.~Nauck and R.~Kruse, ``A neuro-fuzzy method to learn fuzzy classification
  rules from data,'' \emph{Fuzzy sets and Systems}, vol.~89, no.~3, pp.
  277--288, 1997.

\bibitem{de2005elicitation}
M.~De~Cock, C.~Cornelis, and E.~E. Kerre, ``Elicitation of fuzzy association
  rules from positive and negative examples,'' \emph{Fuzzy Sets and Systems},
  vol. 149, no.~1, pp. 73--85, 2005.

\bibitem{hu2003elicitation}
Y.-C. Hu and G.-H. Tzeng, ``Elicitation of classification rules by fuzzy data
  mining,'' \emph{Engineering Applications of Artificial Intelligence},
  vol.~16, no.~7, pp. 709--716, 2003.

\bibitem{roubos2003learning}
J.~A. Roubos, M.~Setnes, and J.~Abonyi, ``Learning fuzzy classification rules
  from labeled data,'' \emph{Information Sciences}, vol. 150, no.~1, pp.
  77--93, 2003.

\bibitem{tang2011fuzzy}
M.~Tang, X.~Chen, W.~Hu, and W.~Yu, ``A fuzzy rule-based classification system
  using interval type-2 fuzzy sets,'' in \emph{Integrated Uncertainty in
  Knowledge Modelling and Decision Making}.\hskip 1em plus 0.5em minus
  0.4em\relax Springer, 2011, pp. 72--80.

\bibitem{wang1992generating}
L.-X. Wang and J.~M. Mendel, ``Generating fuzzy rules by learning from
  examples,'' \emph{Systems, Man and Cybernetics, IEEE Transactions on},
  vol.~22, no.~6, pp. 1414--1427, 1992.

\bibitem{ittner1996discovery}
A.~Ittner and M.~Schlosser, ``Discovery of relevant new features by generating
  non-linear decision trees.'' in \emph{KDD}, 1996, pp. 108--113.

\bibitem{min2009deep}
R.~Min, D.~Stanley, Z.~Yuan, A.~Bonner, Z.~Zhang \emph{et~al.}, ``A deep
  non-linear feature mapping for large-margin knn classification,'' in
  \emph{Data Mining, 2009. ICDM'09. Ninth IEEE International Conference
  on}.\hskip 1em plus 0.5em minus 0.4em\relax IEEE, 2009, pp. 357--366.

\bibitem{khan2001classification}
J.~Khan, J.~S. Wei, M.~Ringner, L.~H. Saal, M.~Ladanyi, F.~Westermann,
  F.~Berthold, M.~Schwab, C.~R. Antonescu, C.~Peterson \emph{et~al.},
  ``Classification and diagnostic prediction of cancers using gene expression
  profiling and artificial neural networks,'' \emph{Nature medicine}, vol.~7,
  no.~6, pp. 673--679, 2001.

\bibitem{burges1998tutorial}
C.~J. Burges, ``A tutorial on support vector machines for pattern
  recognition,'' \emph{Data mining and knowledge discovery}, vol.~2, no.~2, pp.
  121--167, 1998.

\bibitem{chiu1994fuzzy}
S.~L. Chiu, ``Fuzzy model identification based on cluster estimation.''
  \emph{Journal of intelligent and Fuzzy systems}, vol.~2, no.~3, pp. 267--278,
  1994.

\bibitem{kohonen1982self}
T.~Kohonen, ``Self-organized formation of topologically correct feature maps,''
  \emph{Biological cybernetics}, vol.~43, no.~1, pp. 59--69, 1982.

\bibitem{yager1994generation}
R.~R. Yager and D.~P. Filev, ``Generation of fuzzy rules by mountain
  clustering,'' \emph{Journal of Intelligent \& Fuzzy Systems: Applications in
  Engineering and Technology}, vol.~2, no.~3, pp. 209--219, 1994.

\bibitem{zadeh1965fuzzy}
L.~A. Zadeh, ``Fuzzy sets,'' \emph{Information and control}, vol.~8, no.~3, pp.
  338--353, 1965.

\bibitem{zadeh1975concept}
------, ``The concept of a linguistic variable and its application to
  approximate reasoning—i,'' \emph{Information sciences}, vol.~8, no.~3, pp.
  199--249, 1975.

\bibitem{mendel2003type}
J.~M. Mendel, ``Type-2 fuzzy sets: some questions and answers,'' \emph{IEEE
  Connections, Newsletter of the IEEE Neural Networks Society}, vol.~1, pp.
  10--13, 2003.

\bibitem{mendel2001uncertain}
------, \emph{Uncertain rule-based fuzzy logic system: introduction and new
  directions}.\hskip 1em plus 0.5em minus 0.4em\relax Prentice--Hall PTR, 2001.

\bibitem{dyckhoff1984generalized}
H.~Dyckhoff and W.~Pedrycz, ``Generalized means as model of compensative
  connectives,'' \emph{Fuzzy sets and Systems}, vol.~14, no.~2, pp. 143--154,
  1984.

\bibitem{pal1995cluster}
N.~R. Pal and J.~C. Bezdek, ``On cluster validity for the fuzzy c-means
  model,'' \emph{Fuzzy Systems, IEEE Transactions on}, vol.~3, no.~3, pp.
  370--379, 1995.

\bibitem{jin2000fuzzy}
Y.~Jin, ``Fuzzy modeling of high-dimensional systems: complexity reduction and
  interpretability improvement,'' \emph{Fuzzy Systems, IEEE Transactions on},
  vol.~8, no.~2, pp. 212--221, 2000.

\bibitem{jang1991fuzzy}
J.-S.~R. Jang \emph{et~al.}, ``Fuzzy modeling using generalized neural networks
  and kalman filter algorithm.'' in \emph{AAAI}, vol.~91, 1991, pp. 762--767.

\bibitem{jang1993anfis}
J.-S.~R. Jang, ``Anfis: adaptive-network-based fuzzy inference system,''
  \emph{Systems, Man and Cybernetics, IEEE Transactions on}, vol.~23, no.~3,
  pp. 665--685, 1993.

\end{thebibliography}

\end{document}